\pdfoutput=1

\documentclass[11pt]{article}

\usepackage[preprint]{acl}

\usepackage{times}
\usepackage{latexsym}

\usepackage[T1]{fontenc}

\usepackage[utf8]{inputenc}

\usepackage{microtype}

\usepackage{inconsolata}
\usepackage{booktabs}
\usepackage{array}
\usepackage{makecell}
\usepackage{float}
\usepackage{bm}
\usepackage{booktabs}
\usepackage{multirow}
\usepackage{xcolor}
\usepackage{colortbl}
\usepackage{xspace}
\usepackage{subfig}
\usepackage{graphicx} 
\usepackage{color}
\usepackage{paralist}
\usepackage[most]{tcolorbox}

\usepackage{amsmath,amsfonts,amsthm}
\theoremstyle{definition}

\newcommand{\themodel}{CoH\xspace}

\title{Chain-of-History Reasoning for Temporal Knowledge Graph Forecasting}

\author{\textbf{ Yuwei Xia$^{1,2}$\footnotemark[1] , Ding Wang$^{3,4}$\footnotemark[1], Qiang Liu$^{3,4}$\footnotemark[2] ,  Liang Wang$^{3,4}$, Shu Wu$^{3,4}$, Xiaoyu Zhang$^{1,2}$\footnotemark[2]} \\
$^1$Institute of Information Engineering, Chinese Academy of Sciences \\
  $^2$School of Cyber Security, University of Chinese Academy of Sciences \\ 
  $^3$New Laboratory of Pattern Recognition (NLPR), 
        \\ State Key Laboratory of Multimodal Artificial Intelligence Systems, \\ Institute of Automation, Chinese Academy of Sciences \\ 
$^4$School of Artificial Intelligence, University of Chinese Academy of Sciences \\
\texttt{xiayuwei@iie.ac.cn, wangding2024@ia.ac.cn,  qiang.liu@nlpr.ia.ac.cn}\\
  \texttt{liang.wang@cripac.ia.ac.cn, shu.wu@nlpr.ia.ac.cn, zhangxiaoyu@iie.ac.cn} \\
 }

\begin{document}
\maketitle
\renewcommand{\thefootnote}{\fnsymbol{footnote}}
\footnotetext[1]{These authors contributed to the work equally.} 
\footnotetext[2]{To whom correspondence should be addressed.} 
\renewcommand{\thefootnote}{\arabic{footnote}}

\begin{abstract}
Temporal Knowledge Graph (TKG) forecasting aims to predict future facts based on given histories. Most recent graph-based models excel at capturing structural information within TKGs but lack semantic comprehension abilities. Nowadays, with the surge of LLMs, the LLM-based TKG prediction model has emerged. However, the existing LLM-based model exhibits three shortcomings: (1) It only focuses on the first-order history for prediction while ignoring high-order historical information, resulting in the provided information for LLMs being extremely limited. (2) LLMs struggle with optimal reasoning performance under heavy historical information loads. (3) For TKG prediction, the temporal reasoning capability of LLM alone is limited. To address the first two challenges, we propose Chain-of-History (CoH) reasoning which explores high-order histories step-by-step, achieving effective utilization of high-order historical information for LLMs on TKG prediction. To address the third issue, we design \themodel as a plug-and-play module to enhance the performance of graph-based models for TKG prediction. Extensive experiments on three datasets and backbones demonstrate the effectiveness of \themodel.
\end{abstract}

\section{Introduction}
As a carrier of facts with temporal information, Temporal Knowledge Graphs (TKGs) hold significant practical value across various applications \citep{xiang2022temporal, chen2023multi}. Most advanced research on TKGs mainly focuses on predicting future facts occur at time $t_n$ based on given historical facts occur at time $t$ with $t<t_n$. 


Recent supervised methods \citep{2020Recurrent, DBLP:conf/sigir/LiJLGGSWC21, DBLP:conf/ijcai/LiS022} primarily rely on Graph Neural Networks (GNNs) to capture structural dependencies within TKGs, yet they often fall short in effectively modeling semantic information. With the advent of Large Language Models (LLMs), the potential for enhanced temporal reasoning across various tasks is becoming increasingly evident \citep{jain2023language, yuan2023back}. Lee et al. made the first attempt at TKG reasoning using LLMs \citep{lee2023temporal}, presenting histories to LLMs in textual form. Despite these advancements, we contend that significant challenges remain unaddressed.

\begin{figure}
  \centering
  \includegraphics[width=\linewidth]{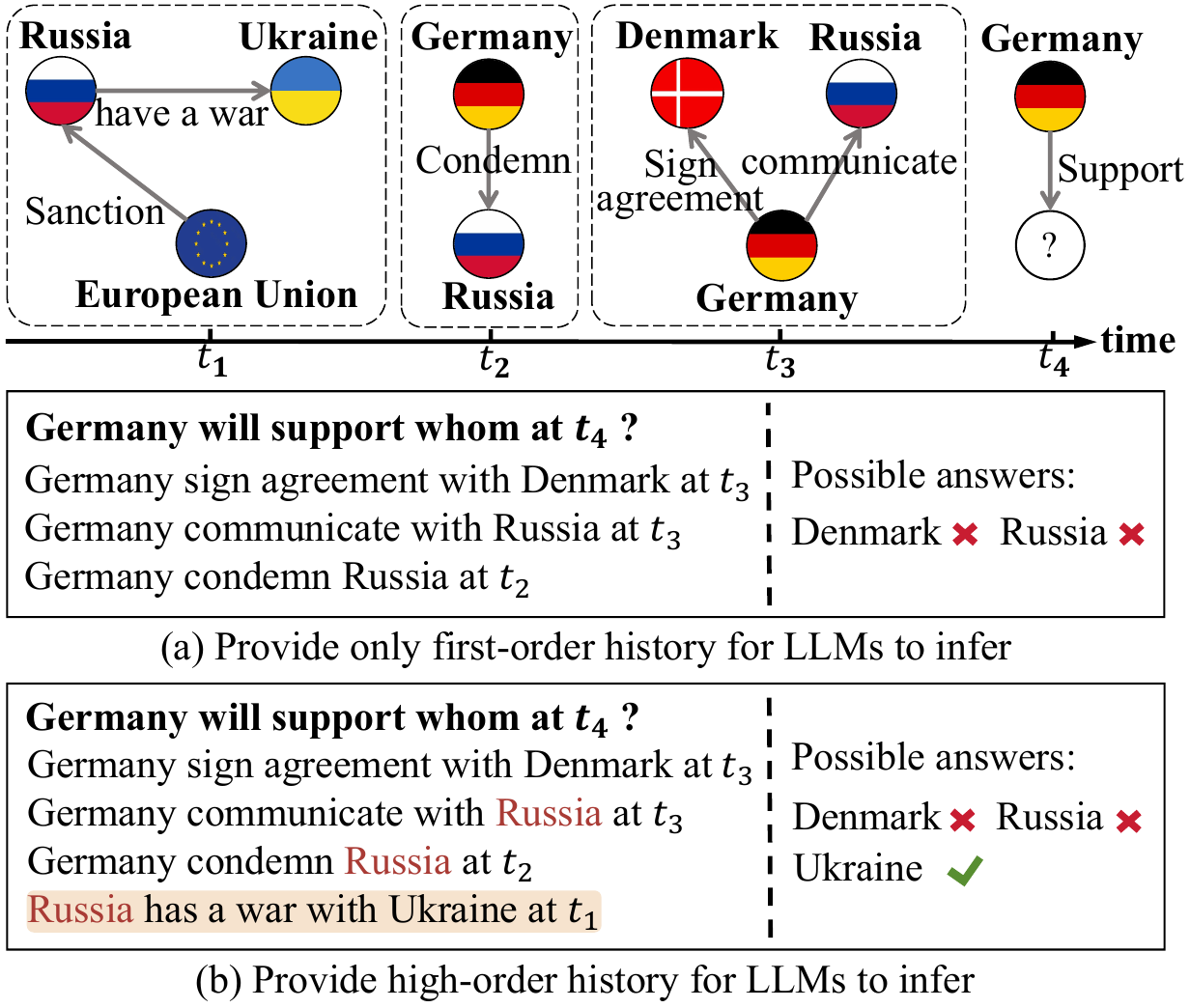}
  \caption{An example of reasoning over TKG with LLMs. In Figures (a) and (b), we provide LLMs with different histories, which prompt LLMs to reason different answers for the predicted fact.}
  \label{mov}
\end{figure}
 
Firstly, the existing TKG prediction model with LLMs only focuses on the first-order histories, ignoring important high-order historical information. Taking Figure \ref{mov} as an example, LLMs aim to infer ``\emph{Germany will support whom at $t_4$}'' with provided histories. The existing model provides LLMs with only first-order histories. In this case, LLMs are constrained to infer wrong answers to ``\emph{Denmark}'' and ``\emph{Russia}'' because given histories fail to encompass the correct answer. When supplied with more high-order histories, LLMs can utilize the history chain ``\emph{Germany$\rightarrow$Russia$\rightarrow$Ukraine}'' to reason the correct answer ``\emph{Ukraine}'' more possibly. 

\begin{figure}
  \centering
  \includegraphics[scale=0.8]{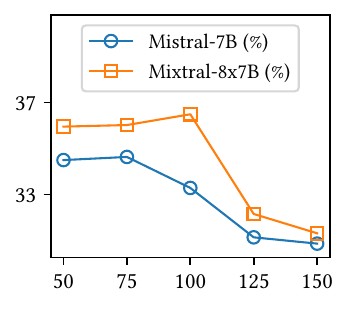}
  \caption{The performance (MRR (\%)) of LLMs of two sizes based on different history lengths on TKG prediction. The provided histories contain both first- and second-order histories. The y-axis represents the MRR (\%) value, and the x-axis denotes the total length of provided first- and second-order histories. The results are based on the commonly used TKG dataset ICEWS14.}
  \label{mov2}
\end{figure}

Secondly, LLMs struggle to maintain reasoning performance under heavy historical information loads. Intuitively, we provide more comprehensive high-order histories for LLMs to infer. However, as shown in Figure \ref{mov2}, the performance of LLMs does not necessarily improve or remain stable with the increase in history length, instead experiencing a steep decline beyond a certain threshold of history length regardless of the model size. This indicates that over-complicated historical information may confuse LLMs \cite{shi2023language}, making LLMs hard to reason correct answers. Thus, exploring ways to offer higher-order histories for LLMs effectively is a worthwhile investigation.

Thirdly, relying solely on the reasoning capabilities of LLMs still remains limited on TKG prediction. Though LLMs possess unique semantic comprehension advantages in reasoning, they still struggle to achieve the same level of ability in capturing complex structural information as graph-based models. However, this unique advantage of LLMs precisely compensates for the shortcomings of graph-based models in modeling semantic information, thereby enhancing the performance of graph-based models on TKG prediction. 

\begin{figure*}
  \centering
  \includegraphics[width=\linewidth]{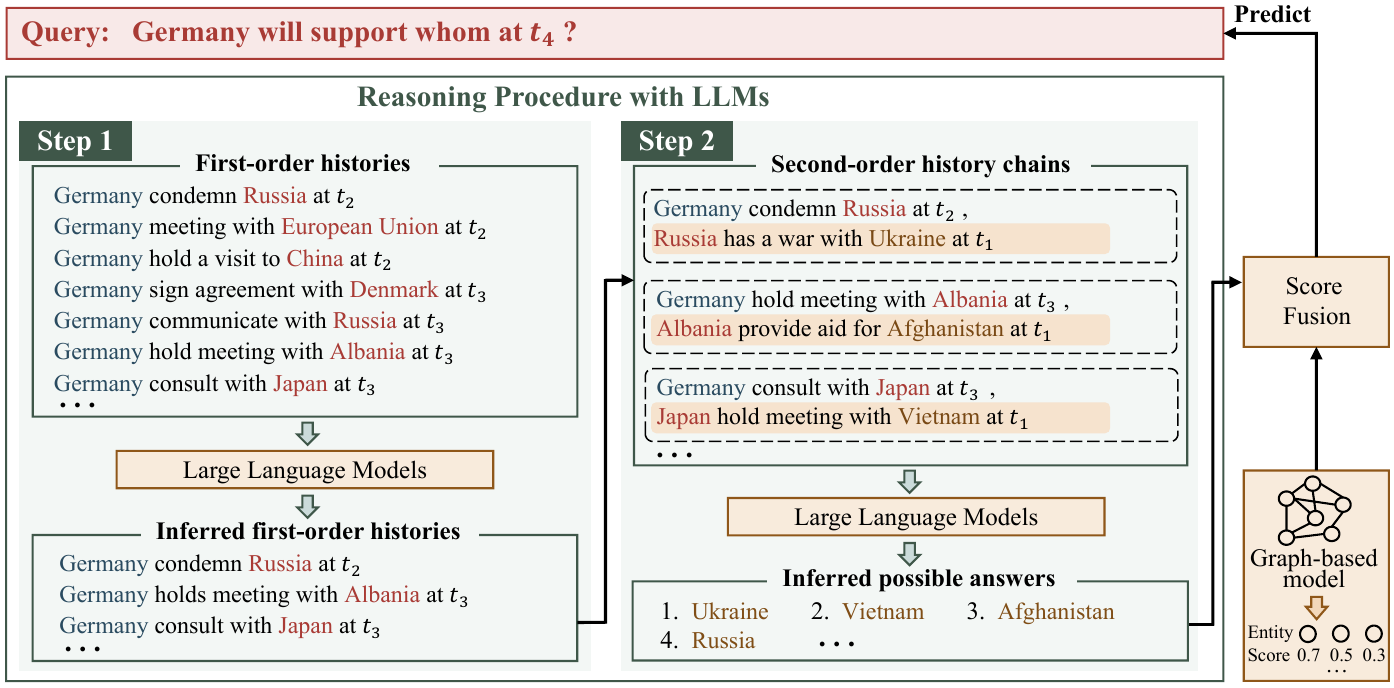}
  \caption{
  An illustration of a two-step \themodel reasoning procedure. In the first step, LLMs are provided with only first-order histories and asked to infer the most important histories. In the second step, LLMs are provided with second-order history chains based on the inferred first-order histories and asked to infer possible answers to the given query. Then the answers inferred by LLMs and graph-based models are adaptively fused to make the final prediction. \textbf{Note that this only serves as a two-step reasoning example, more steps can be executed with \themodel.}
 }
 \label{framework}
\end{figure*}

To address the above issues, we propose a \underline{C}hain-\underline{o}f-\underline{H}istory (CoH) reasoning method for TKG prediction. Instead of providing LLMs with all histories at once, \themodel provides LLMs with high-order histories step-by-step. Specifically, \themodel adopts LLMs to explore important high-order history chains step-by-step, and reason the answers to the query only based on inferred history chains in the last step. A two-step \themodel reasoning procedure is shown in Figure \ref{framework}. This also can extend to multiple-step reasoning for complex situations. LLMs can continue inferring important second-order history chains in Step 2, then infer answers with third-order history chains in Step 3, and so on. In this way, LLMs only need to process a limited quantity of histories at each step, preventing an overwhelming influx of complex information while effectively leveraging a more comprehensive set of high-order information. This approach enables LLMs to perform more accurate reasoning with higher-order information for TKG prediction.
Moreover, we design \themodel as a plug-and-play module for TKG reasoning. As shown in Figure \ref{framework}, we fuse the predicted results obtained by LLMs and graph-based TKG models to make the final prediction more comprehensively. To summarize, the contributions of this paper can be listed as follows:
\begin{itemize}
    \item We are the first to explore the necessity and difficulty of providing numerous high-order histories for LLMs on TKG prediction. And we propose the Chain-of-History reasoning method which adopts LLMs to explore the history chains step-by-step.
    \item We are the first to propose enhancing the performance of graph-based TKG models with LLMs, utilizing the semantic understanding advantage of LLM to compensate for the shortcoming of graph-based models.
    \item We conduct extensive experiments on three commonly used TKG datasets and three graph-based TKG models, the results demonstrate the effectiveness of \themodel.
\end{itemize}

\section{Problem Formulation} 
\paragraph{Temporal Knowledge Graph Prediction.} Let $\mathcal{E}$ and $\mathcal{R}$ represent a set of entities and relations. A Temporal Knowledge Graph (TKG) $\mathcal{G}$ can be defined as $\mathcal{G} = \{\mathcal{G}_1, \mathcal{G}_2, \cdots, \mathcal{G}_n\}$. Each $\mathcal{G}_t \in \mathcal{G}$ contains facts that occur at time $t$. Each fact is represented as a quadruple $(s, r, o, t)$, in which $s, o \in \mathcal{E}$ and $r \in \mathcal{R}$. Given a query $q=(s^q, r^q, ?, t^q)$ or $q=(?, r^q, o^q, t^q)$, TKG prediction task aims to predict the missing object entity or subject entity with historical KG sequence $\mathcal{G}_{<t^q}=\{\mathcal{G}_1, \mathcal{G}_2, \cdots, \mathcal{G}_{t^q-1}\}$. The candidate answers for $q$ are all entities in $\mathcal{E}$, each candidate $e_i \in \mathcal{E}$ will be estimated with a score by TKG prediction models. 

\paragraph{High-order History Chains in TKGs.} For each query $q=(s^q, r^q, ?, t^q)$ to be predicted, we denote $\{(s^q, r, o, t)|(s^q, r, o, t)\in\mathcal{G}_{<t^q}\}$ as the first-order histories of $q$. If $(s^q, r, o, t)$ is the first-order history of $q$, we denote histories in the form of $(o, r, o', t)$ as a set of second-order histories of $q$. The higher-order histories of $q$ are deduced in this way. And we denote $[(s^q, r, o, t),(o, r, o', t^j)]$ as a second-order history chain of $q$, which consists of a first-order history of $q$ and its associated second-order history of $q$. The higher-order history chains of $q$ can be deduced in this way.

\section{Chain-of-History Reasoning over Temporal Knowledge Graph}
For a given query $q=(s^q, r^q, ?, t^q)$, \themodel predicts the answers by exploring history chains related to $q$ step-by-step, then answers are fused with predicted results by graph-based TKG models to make the final prediction for $q$. In this section, we illustrate \themodel shown in Figure \ref{framework} in detail. Section \ref{History Processing} explains how to convert quadruples in TKGs into text formats suitable for LLMs. Section \ref{Chain-of-History Reasoning with LLMs} demonstrates how to properly provide histories from $\mathcal{G}_{<t^q}$ for LLMs and how to instruct LLMs to reason in each step. Section \ref{Predicted Results Fusion} shows how to transform answers predicted by LLMs into scores and fuse them with the results from graph-based TKG models.

\subsection{History Processing}
\label{History Processing}
Each fact in TKGs is presented as a quadruple $(s, r, o, t)$, such as ``(Germany, Sign agreement, Denmark, 2023-06-02)''. To make each quadruple more linguistically comprehensible for LLMs, we introduce prepositions to transform the quadruple into a more fluent sentence. Most importantly, considering the prior knowledge of LLMs, we process the time ``2023-06-02'' in each quadruple into a more abstract form like ``153rd day'' to prevent LLMs from directly exploiting prior knowledge for predictions. Consequently, we provide LLMs with each quadruple in the form of ``(Germany Sign agreement with Denmark on the 153rd day)''.

\begin{table*}[!ht]
	\centering
	\resizebox{\linewidth}{!}
	{\begin{tabular}{l|p{16cm}}
			\toprule
\textbf{Step $i$}& \textbf{Instruction}\\
                \midrule
Step 1 &
There is a given text consisting of multiple historical events in the form of ``\{id\}:[\{subject\} \{relation\} \{object\} \{time\}];''. And there is a query in the form of: ``\{subject\} \{relation\} \{whom\} time\}?'' If you must infer several \{object\} that you think may be the answer to the given query based on the given historical events, \textcolor[RGB]{168,114,9}{what important historical events do you base your predictions on? Please list the top $n$ most important histories} and output their \{id\}.\\
			\midrule
Step 2 to Step $k$-1&
There is a given text consisting of multiple history chains in the form of ``\{id\}:[\{subject\} \{relation\} \{object\} \{time\}, \{subject\} \{relation\} \{object\} \{time\}, ...];''. And there is a query in the form of: ``\{subject\} \{relation\} \{whom\} time\}?'' If you must infer several \{object\} that you think may be the answer to the given query based on the given historical events, \textcolor[RGB]{16,99,236}{what important history chains do you base your predictions on? Please list the top $n$ most important history chains} and output their \{id\}.\\
\midrule
Step $k$&
You must be able to correctly predict the \{whom\} of the given query from a given text consisting of multiple historical events in the form of ``\{subject\} \{relation\} \{object\} \{time\}'' and the query in the form of ``\{subject\} \{relation\} \{whom\} \{time\}?'' \textcolor[RGB]{243,67,52}{You must output several \{object\} that you think may be the answer to the given query based on the given historical events.} Please list all possible \{object\} which may be answers to the query. Please assign each answer a serial number to represent its probability of being the correct answer. Note that answers with a high probability of being correct should be listed first.\\
			\bottomrule
	\end{tabular}}
        \caption{Instruction design for each step in a $k$-step \themodel reasoning procedure.}
	\label{prompt}
\end{table*}

\subsection{Reasoning Steps}
\label{Chain-of-History Reasoning with LLMs}
In a $k$-step reasoning procedure of \themodel, LLMs are instructed to explore the most significant history chains related to the given query $q$ from Step 1 to Step $k$-1. Subsequently, LLMs reason the possible answers to $q$ in Step k with k-order history chains.

\paragraph{Step 1 to Step $k$-1 Reasoning.} In Step 1, LLMs are provided with only first-order histories of $q$, and are instructed to reason $n$ first-order histories that mostly contribute to answering $q$. From Step 2 to Step $k$-1, LLMs are provided with $i$-order history chains in Step $i$ ($i=\{2,3,\cdots,k-1\}$), and are instructed to infer $n$ most significant history chains. The instruction design is shown in Table \ref{prompt}. Within this sequence of steps, the outputs of LLMs in Step $i$-1 are the inferred ($i$-1)-order history chains, then each of which is supplied with corresponding $i$-order histories to consist of $i$-order history chains. These history chains subsequently serve as input for the next Step $i$. As shown in Figure \ref{framework}, ``\emph{Germany condemn Russia at $t_2$}'' is one of the outputs in Step 1. Then it is supplied with corresponding second-order history ``\emph{Russia has a war with Ukraine at $t_1$}'' to consist of the second-order history chain, which serves as the input for Step 2. The prompt example for Step 1 and Step $i$ ($i=\{2,3,\cdots,k-1\}$) can refer to Appendix \ref{Prompt Example}.

\paragraph{Step $k$ Reasoning.} In Step $k$, LLMs are provided with $k$-order history chains and instructed to reason possible answers for the given query $q$. The instruction design for Step $k$ is shown in Table \ref{prompt}. Especially, we instruct LLMs to prioritize outputting the entity with a higher probability of being the correct answer. As shown in Figure \ref{framework}, the output of Step 2 includes several possible answers to $q$. Each answer is assigned a numerical index (1,2,3,$\cdots$), with a lower index indicating a higher probability of the answer being correct. The prompt example for Step $k$ can refer to Appendix \ref{Prompt Example}.

\subsection{Results Processing and Fusion}
\label{Predicted Results Fusion}
In graph-based TKG models, each entity and relation in a quadruple are denoted with an id like ``(30, 13, 8, 2023-06-02)'' instead of ``(Germany, Sign agreement, Denmark, 2023-06-02)''. The lack of semantic modeling of histories makes graph-based models mainly depend on structural information within TKGs for prediction. However, entities and relations inherently carry semantic information, which also constitutes a significant part of TKGs. Considering the importance of both structural and semantic information within TKGs, we propose to fuse the predicted results of LLMs and graph-based models to obtain more comprehensive results for predicting over TKGs more accurately.

Firstly, for a given $q$, we need to obtain the score of each entity $e_i$ in the LLMs' predicted answer set $\mathcal{A}_{\rm{LLM}}^q$. As we mentioned each answer predicted by LLMs in Step $k$ is assigned an index, which represents the probability of the answer being correct. We convert the index of each answer $e_i\in \mathcal{A}_{\rm{LLM}}^q$ into its corresponding score with an exponential decay function as follows:
\begin{equation}
    \mathcal{S}_{{\rm{LLM}}}^{e_i}=\frac{1}{1 + e^{\alpha \cdot {\rm{idx}}^{e_i}}},
\end{equation}
where $\mathcal{S}_{\rm{LLM}}^{e_i}$ denotes the score of the entity $e_i$ obtained with LLMs for being the answer to $q$, ${\rm{idx}}^{e_i}$ represents the numerical index of the answer $e_i$ in the outputs of LLMs, and $\alpha$ is a hyper-parameter to control the score disparity among answers with different indexes. Note that since the outputs of LLMs can not include all candidate entities in $\mathcal{E}$ like graph-based models, we assign the score of $e_i$ as $0$ where $e_i\in\mathcal{E}$ but $e_i\notin \mathcal{A}_{\rm{LLM}}^q$.

Then, we can fuse the score of each candidate entity $e_i\in\mathcal{E}$ obtained with LLMs and graph-based models as follows:
\begin{equation}
    \mathcal{S}^{e_i}=w\cdot \mathcal{S}_{{\rm{Graph}}}^{e_i}+(1-w)\cdot \mathcal{S}_{{\rm{LLM}}}^{e_i},
\end{equation}
where $\mathcal{S}_{{\rm{Graph}}}^{e_i}$ denotes the score of $e_i$ obtained with graph-based models, and $w$ is a hyper-parameter to determine the weight of different scores. $\mathcal{S}^{e_i}$ represents the comprehensive score of the candidate $e_i$. Finally, the ranked candidate list based on comprehensive scores is used for predicting $q$.

\section{Experiments}
In this section, we conduct extensive experiments to evaluate and analyze \themodel on three typical datasets and three backbones for TKG prediction. Details of datasets and backbones can be referred to Appendix \ref{Datasets} and \ref{backbones}, respectively. 

\subsection{Experimental Settings}
\subsubsection{Evaluation}
For evaluation, we adopt widely used metrics MRR and Hits@\{1, 3, 10\} in experiments. Without loss of generality \cite{DBLP:conf/sigir/LiJLGGSWC21}, we only report the experimental results under the raw setting. Note that different from \citep{lee2023temporal}, we fully align the evaluation mechanism for LLMs in TKG prediction with those used in graph-based models to ensure a more fair comparison. Specifically, 
during the testing phase of graph-based models, the test set is typically augmented by doubling its size through reversing $(s, r, o, t)$ into $(o, r^{-1}, s, t)$, to assess the model's performance more comprehensively. Correspondingly, we also evaluate LLMs on TKG prediction with reversed test sets.

\subsubsection{\themodel Implementation Details}
In this paper, we implement \themodel with two-step reasoning based on an open-sourced language model Mixtral-8x7B \citep{jiang2024mixtral}. In Step 1, we provide the LLM with 100 first-order histories and set $n$ to 30, allowing the LLM to infer the most important 30 first-order histories from the given ones. In Step 2, we do not strictly limit the number of answers output from the LLM. For more details on implementation please refer to Appendix \ref{Implementation Details}.

\begin{table*}[!ht]
    \centering
    \resizebox{\textwidth}{!}{
    \begin{tabular}{c|c|cccccccccccc}
    \toprule
    \multirow{2.5}[0]*{\textbf{Model Type}} &\multirow{2.5}[0]{*}{\textbf{Model}} &\multicolumn{4}{c}{\textbf{ICEWS14}} & \multicolumn{4}{c}{\textbf{ICEWS18}}  & \multicolumn{4}{c}{\textbf{ICEWS05-15}} \\
    \cmidrule(lr){3-6} \cmidrule(lr){7-10} \cmidrule(lr){11-14}
    ~ &~ &MRR &  Hit@1 &  Hit@3& Hit@10 &MRR &Hit@1 &  Hit@3 & Hit@10 & MRR &Hit@1 &  Hit@3 & Hit@10 \\
    \midrule 
    \multirow{3}*{LLMs} &ICL \citep{lee2023temporal}* &31.79	&22.38	&37.67	&47.70	&21.51	&14.77	&26.08	&40.57	&35.34	&25.18	&43.92	&56.24 \\
    ~ &\themodel &\textbf{34.51}	&\textbf{24.20}	&\textbf{39.67}	&\textbf{51.21}	&\textbf{23.94}	&\textbf{16.81}	&\textbf{28.15}	&\textbf{42.68}	&\textbf{37.51}	&\textbf{27.72}	&\textbf{47.17}	&\textbf{59.58} \\
    \rowcolor{gray!15} \cellcolor{white}~ &$\Delta$\emph{Improve} &8.56\%	&8.13\%	&5.31\%	&7.36\%	&11.30\%	&13.81\%	&7.94\%	&5.20\%	&6.14\%	&10.09\%	&7.40\%	&5.94\% \\
    \midrule
    \multirow{15}*{LLMs + Graph} &RE-NET &38.75 	&28.96 	&43.64 	&57.61 	&28.72 	&18.84 	&32.66 	&48.18 	&44.05 	&33.22 	&51.23 	&65.02 \\
    ~ & RE-NET + ICL* &39.39	&29.12	&44.37	&58.25	&29.01	&18.98	&33.11	&48.78	&45.12	&33.98	&52.09	&66.23\\
    ~ & RE-NET + \themodel &\textbf{40.43} &\textbf{30.34} &\textbf{45.78} &\textbf{60.42} &\textbf{29.77} &\textbf{19.96} &\textbf{34.14} &\textbf{49.59} &\textbf{46.37} &\textbf{34.99} &\textbf{53.13} &\textbf{67.71} \\
    \rowcolor{gray!15} \cellcolor{white}~ &$\Delta$\emph{Improve}* &1.65\%	&0.56\%	&1.67\%	&1.11\%	&1.00\%	&0.73\%	&1.39\%	&1.25\% &2.42\%	&2.29\%	&1.67\%	&1.87\%\\
    \rowcolor{gray!15} \cellcolor{white}~ &$\Delta$\emph{Improve}  &4.34\% &4.77\%	&4.90\%	&4.87\%	&3.66\%	&5.94\%	&4.53\%	&2.93\%	&5.26\%	&5.33\%	&3.71\%	&4.14\% \\
    \cmidrule{2-14}
    ~ & RE-GCN &41.33	&30.61	&46.66	&62.31	&31.08	&20.44	&35.39	&52.06	&46.89	&35.5	&53.33	&68.4 \\
    ~ &  RE-GCN + ICL* &41.84	&30.84	&47.27	&62.97	&31.31	&20.65	&35.79	&52.61	&47.87	&36.16	&54.24	&69.43\\
    ~ & RE-GCN + \themodel &\textbf{42.41} 	&\textbf{31.77} 	&\textbf{47.85} &\textbf{63.80} 	&\textbf{32.10} 	&\textbf{21.75} 	&\textbf{36.51} 	&\textbf{53.37} 	&\textbf{47.98} 	&\textbf{37.53} 	&\textbf{54.94} 	&\textbf{70.68} \\
    \rowcolor{gray!15} \cellcolor{white}~ &$\Delta$\emph{Improve}* & 1.23\%	&0.75\%	&1.31\%	&1.06\%	&0.74\%	&1.03\%	&1.13\%	&1.06\%	&2.09\%	&1.86\%	&1.71\%	&1.51\% \\
    \rowcolor{gray!15} \cellcolor{white}~ &$\Delta$\emph{Improve} &2.61\%	&3.79\%	&2.55\%	&2.39\%	&3.28\%	&6.41\%	&3.16\%	&2.52\%	&2.32\%	&5.72\%	&3.02\%	&3.33\% \\
    \cmidrule{2-14} 
    ~ & TiRGN &42.93	&32.1	&48.53	&63.6	&31.97	&20.95	&36.67	&53.66	&48.5	&36.87	&55.19	&70.27\\
    ~ & TiRGN + ICL* &43.27	&32.28	&49.04	&64.17	&32.18	&21.07	&36.98	&54.04	&49.15	&37.25	&55.73	&70.9\\
    ~ & TiRGN + \themodel &\textbf{43.94}	&\textbf{33.07}	&\textbf{49.64}	&\textbf{64.90}	&\textbf{32.98}	&\textbf{21.83}	&\textbf{37.79}	&\textbf{54.92}	&\textbf{49.71}	&\textbf{38.01}	&\textbf{56.40} &\textbf{71.25} \\
    \rowcolor{gray!15} \cellcolor{white}~ &$\Delta$\emph{Improve}* &0.79\%	&0.56\%	&1.05\%	&0.90\%	&0.66\%	&0.57\%	&0.85\%	&0.71\%	&1.34\%	&1.03\%	&0.98\%	&0.90\%\\
    \rowcolor{gray!15} \cellcolor{white}~ &$\Delta$\emph{Improve} &2.35\%	&3.02\%	&2.29\%	&2.04\%	&3.16\%	&4.20\%	&3.05\%	&2.35\%	&2.49\%	&3.09\%	&2.19\%	&1.39\% \\
    \bottomrule
    \end{tabular}  
    }
    \caption{Performance comparison of \themodel on TKG prediction on three datasets in terms of MRR (\%), Hit@1 (\%), Hit@3 (\%), and Hit@10 (\%). All results are obtained under raw metrics. The highest performance is highlighted in bold. And * represents the reproduced model with the same evaluation and LLM used in this paper. $\Delta$\emph{Improve} and $\Delta$\emph{Improve}* indicate the relative improvements of \themodel and the ICL-based model plugged into the graph-based models over the original graph-based backbones in percentage, respectively.}
    \label{main performance}
\end{table*}

\subsection{Performance Comparison}
\label{Performance Comparison (RQ1)}
In this section, we present a comprehensive evaluation of the proposed \themodel. We first evaluate the performance of only utilizing LLMs on TKG prediction with \themodel reasoning. Then we plug \themodel on three existing state-of-the-art graph-based TKG prediction models to see the potential gains it can yield. The results are shown in Table \ref{main performance}, from which we have the following observations.

On the one hand, from the results of LLMs, the two-step \themodel reasoning outperforms ICL \citep{lee2023temporal} which solely provides LLMs with first-order histories under all evaluation metrics on three datasets. This indicates the usefulness of the higher-order histories provided step-by-step. And we observe that the relative improvements of \themodel over the existing method are more obvious on ICEWS18 than other datasets, which implies that ICEWS18 may contain more and complex information of history chains. Despite the progress achieved by \themodel, the performance of only utilizing LLMs on TKG prediction is still pretty limited compared with graph-based models.

On the other hand, though the temporal reasoning capability of LLMs on TKG prediction is relatively limited, they can be flexibly used as a plug-and-play module to enhance the performance of graph-based models. From the results of plugging \themodel and the ICL-based model into existing graph-based models, we can see that the two LLM-based models can effectively improve their performance. The gains introduced by \themodel to the performance of graph-based models far surpass those achieved by the ICL-based model, which further demonstrates the effectiveness of our proposed model. Moreover, we analyze the rationale behind these gains may be attributed to the distinct reasoning mechanisms of LLMs and graph-based models, each of which possesses unique strengths. In this case, the powerful semantic understanding ability of LLMs may be capable of compensating to some extent for the inherent limitations in semantic information modeling of graph-based models.

\begin{table*}[!ht]
    \centering
    \resizebox{0.9\textwidth}{!}{
    \begin{tabular}{c|c|cccccccccccc}
    \toprule
    \multirow{2.5}[0]*{\textbf{Model}} &\multirow{2.5}[0]{*}{\textbf{Step $i$}} &\multicolumn{4}{c}{\textbf{ICEWS14}} & \multicolumn{4}{c}{\textbf{ICEWS18}}  & \multicolumn{4}{c}{\textbf{ICEWS05-15}} \\
    \cmidrule(lr){3-6} \cmidrule(lr){7-10} \cmidrule(lr){11-14}
    ~ &~ &MRR &  Hit@1 &  Hit@3& Hit@10 &MRR &Hit@1 &  Hit@3 & Hit@10 & MRR &Hit@1 &  Hit@3 & Hit@10 \\
    \midrule 
    \multirow{2}*{\themodel w/o LR} & Step 1 &32.31	&23.18	&37.92	&49.15	&20.97	&13.88	&24.83	&36.52	&34.5	&22.24	&42.01	&56.18\\
    ~ &Step 2 &32.68	&23.74	&38.12	&50.09	&21.89	&14.17	&25.32	&37.98	&34.89	&22.48	&43.26	&57.34\\
    \midrule
    \multirow{2}*{\themodel} & Step 1 &33.97	&23.86	&39.03	&49.96	&22.03	&15.34	&26.82	&40.57	&36.48	&24.84	&46.44	&58.51\\
    ~ &Step 2 &\textbf{34.51}	&\textbf{24.20}	&\textbf{39.67}	&\textbf{51.21}	&\textbf{23.94}	&\textbf{16.81}	&\textbf{28.15}	&\textbf{42.68}	&\textbf{37.51}	&\textbf{27.72}	&\textbf{47.17}	&\textbf{59.58}\\
    \midrule
    \themodel w/o IS & Step 2 &24.57	&12.31	&34.75	&51.09	&13.8	&9.14	&20.57	&35.67	&29.61	&18.55	&37.64	&56.87\\
    \bottomrule
    \end{tabular}
    }
    \caption{Ablation studies to investigate the effectiveness of the high-order historical information, step-by-step mechanism, and score ranking procedure of \themodel in terms of MRR (\%), Hit@1 (\%), Hit@3 (\%), and Hit@10 (\%). And all results are obtained under raw metrics.}
    \label{ablation}
\end{table*}

\subsection{Ablation Study}
In this section, we conduct experiments to investigate the effectiveness of the high-order historical information, step-by-step reasoning mechanism, and score ranking procedure in \themodel. The results are shown in Table \ref{ablation}.

\paragraph{Analysis of high-order historical information.} To verify the usefulness of the high-order historical information for TKG prediction with LLMs, we use the inferred first-order histories in Step 1 and the second-order history chains in Step 2 to predict answers, respectively. From the results of \themodel shown in Table \ref{ablation} we can observe that, the predicted results of LLMs based on the second-order history chains are notably superior to the results derived solely from first-order histories. This indicates the usefulness of second-order histories.

\paragraph{Analysis of step-by-step reasoning mechanism.} Specifically, ``LR'' in Table \ref{ablation} denotes the step of reasoning important first-order histories by LLMs in two-step \themodel reasoning. And we implement ``\themodel w/o LR'' by replacing the inferred $n$ first-order histories by LLMs with $n$ first-order histories in the latest timestamps. In this way, we can find out whether LLMs can deduce meaningful historical information within the step-by-step reasoning mechanism. From the results, we can see that \themodel outperforms the one without LR under all evaluation metrics on three datasets, which illustrates the effectiveness of achieving the step-by-step reasoning mechanism with LLMs. 

\paragraph{Analysis of the score ranking procedure.} In the last step of \themodel reasoning, we instruct LLMs to output possible answers in order based on their possibilities of being correct. To verify whether the output index of each answer is related to its correctness, we shuffle the index order of the answers, which is denoted as ``\themodel w/o IS'' in Table \ref{ablation}. Compared the results of \themodel w/o IS with \themodel, we can see that the shuffled index order leads to a huge drop in performance. This indicates that the indexes outputted by LLMs can be helpful for score ranking on TKG prediction.

\begin{table*}[!ht]
	\centering
	\resizebox{\textwidth}{!}{
		\begin{tabular}{l|l}
            \toprule
    \rowcolor{gray!15} \cellcolor{white}\textbf{Model} & \textbf{Query 1: Military\_(Myanmar) Express\_intent\_to\_meet\_or\_negotiate to whom on the 351th day? (GT: Thailand)}\\
    \midrule
    \multirow{7}*\themodel & \textbf{[Military\_(Myanmar) Fight\_with\_small\_arms\_and\_light\_weapons Myanmar 338]}$\boldsymbol{\rightarrow}$ \textcolor[RGB]{16,99,236}{[Myanmar Make\_a\_visit to Thailand 328]}\\
    ~ & \textbf{[Military\_(Myanmar) Charge\_with\_legal\_action Media\_Personnel\_(Myanmar) 305]}$\boldsymbol{\rightarrow}$ \textcolor[RGB]{16,99,236}{[Media\_Personnel\_(Myanmar) Make\_statement Detainee\_(Myanmar) 302]}\\
    ~ & \textbf{[Military\_(Myanmar) Make\_an\_appeal\_or\_request Citizen\_(Thailand) 272]}$\boldsymbol{\rightarrow}$\textcolor[RGB]{16,99,236}{[Citizen\_(Thailand) Use\_violence to Thailand 271]}\\
    ~& \textbf{[Military\_(Myanmar) Make\_an\_appeal\_or\_request Citizen\_(Thailand) 271]} $\boldsymbol{\rightarrow}$\textcolor[RGB]{16,99,236}
 {[Citizen\_(Thailand) Release\_person(s) Activist\_(Thailand) 267]}\\
 ~ &[Military\_(Myanmar) Praise\_or\_endorse Military\_(Thailand) 185]\\
    ~ & [Military\_(Myanmar) Use\_military\_force to Rebel\_Group\_(Myanmar) 174] \\
    ~& $\boldsymbol\cdots$ $\boldsymbol\cdots$\\
    ~ & \textbf{[Military\_(Myanmar) Use\_military\_force to National\_Liberation\_Army 62]}$\boldsymbol{\rightarrow}$\textcolor[RGB]{16,99,236}{[National\_Liberation\_Army Make\_statement to Guerrilla\_(Colombia) 58]}\\
    \cmidrule{2-2}
    ~ & Answer: \textcolor[RGB]{243,67,52}{1. Thailand} $\xspace$ 2. Citizen\_Thailand $\xspace$ 3. Activist\_Thailand $\xspace$ 4. National\_Liberation\_Army $\xspace$ 5. Media\_Personnel\_Myanmar\\
    \midrule
    Graph & Answer: 1. Myanmar $\xspace$ 2. Malaysia $\xspace$ 3. Cambodia $\xspace$ 4. Citizen\_(Thailand) $\xspace$ \textcolor[RGB]{243,67,52}{5. Thailand}\\
    \midrule
    \rowcolor{gray!15} \cellcolor{white}~ & \textbf{Query 2: Saudi\_Army Use\_military\_force to whom on the 3744th day?
 (GT: Armed\_Rebel\_(Yemen))}\\
    \midrule
    \multirow{7}*\themodel & \textbf{[Saudi\_Army Make\_an\_appeal\_or\_request Zillur\_Rahman 2581]} $\boldsymbol{\rightarrow}$\textcolor[RGB]{16,99,236}{[Zillur\_Rahman Make\_empathetic\_comment Citizen\_(North\_Korea) 2543]}\\
~ &\textbf{[Saudi\_Army Consult Zillur\_Rahman 2581]} $\boldsymbol{\rightarrow}$ \textcolor[RGB]{16,99,236}{ [Zillur\_Rahman Make\_empathetic\_comment citizen\_(Nerth\_Kerea) 2543]}\\ 
~ &\textbf{[Saudi\_Army Express\_intent\_to\_meet\_or\_negotiate with Thailand 2581]} $\boldsymbol{\rightarrow}$ \textcolor[RGB]{16,99,236}{[Thailand Use\_military\_force Military\_(Cambodia) 2580]}\\
~ &\textbf{[Saudi\_Army Investigate Armed\_Gang\_(Saudi\_Arabia) 1798]} $\boldsymbol{\rightarrow}$ \textcolor[RGB]{16,99,236}{[Armed\_Gang\_(Saudi\_Arabia) Use\_unconventional\_violence Citizen\_(Saudi\_Arabia) 1773]}\\
~ &\textbf{[Saudi\_Army Employ\_aerial\_weapons Yemen 1769]} $\boldsymbol{\rightarrow}$ \textcolor[RGB]{16,99,236}{[Yemen Charge\_with\_legal\_action Armed\_Rebel\_(Yemen) 1764]}\\

    \cmidrule{2-2}
    ~ & Answer: 1. Military\_(Cambodia) $\xspace$ \textcolor[RGB]{243,67,52}{2. Armed\_Rebel\_(Yemen)} $\xspace$ 3. Armed\_Gang\_(Saudi\_Arabia) $\xspace$ 4. Citizen\_(Saudi\_Arabia) $\xspace$ 5. Citizen\_(North\_Korea) \\
    \midrule
    Graph & Answer: 1. Yemen $\xspace$ 2. Citizen\_(Saudi\_Arabia) $\xspace$ 3. Saudi\_Arabian\_Defence\_Forces $\xspace$ 4. Police\_(Saudi\_Arabia) $\xspace$ \textcolor[RGB]{243,67,52}{5. Armed\_Rebel\_(Yemen)} \\
            \bottomrule
        \end{tabular}	}
	\caption{Case studies with two queries for showing the procedure of \themodel reasoning. The bold histories denote the inferred first-order histories by LLMs in Step 1, and the histories in blue color denote corresponding second-order histories. These two kinds of histories consist of second-order history chains for LLMs to infer answers in Step 2. And the answers in red color represent the ground truth of the given query.}
	\label{case study}
\end{table*}

\subsection{Case Study}
In this section,  we visualize the reasoning process of two queries to understand the reasoning mechanism of \themodel. Moreover, to further understand how \themodel benefits the graph-based TKG prediction models, we show the difference between the results predicted by \themodel and those inferred by the graph-based model RE-NET \citep{2020Recurrent}. Details of the two cases are shown in Table \ref{case study}.

From the reasoning process of the two cases, we can see that LLMs possess the capability of inferring important histories related to the given query. And in the way of \themodel reasoning, LLMs can accurately infer the answers like ``Thailand'' in the second-order history chains. Furthermore, comparing the answers of \themodel with the graph-based model, we can see that \themodel can infer the correct answer more accurately in some scenarios. We analyze the potential reason is that the semantic reasoning capability can allow LLMs to identify crucial historical information.

For example, from the numerous histories in query 1, LLMs can more precisely infer facts involved with relations like ``\emph{Fight with small arms}'', ``\emph{Charge with legal action}'', or ``\emph{Make an appeal}'' that are more likely to result in the occurrence of ``\emph{Express intent to meet or negotiate}'' based on semantic comprehension. For query 2, understanding the semantic meaning differences and correlations between entity ``\emph{Yemen}'' and ``\emph{Armed\_Rebel\_(Yemen)}'' is a very crucial clue. However, the two entities are just regarded as two different IDs in graph-based models, and available histories for query 2 are too limited for them to aggregate abundant information. In this case, the semantic reasoning characteristics of \themodel may be more advantageous. Consequently, LLMs exhibit semantic comprehension capability, while the graph-based models possess a powerful ability to capture structural information. Thus, the reasoning capability of \themodel based on LLMs may potentially complement the reasoning performed by graph-based models in certain scenarios.  

\begin{table*}[!ht]
	\centering
	\resizebox{\textwidth}{!}{
		\begin{tabular}{l|l}
            \toprule
    \rowcolor{gray!15} \cellcolor{white}\textbf{Query} & \textbf{Federica\_Mogherini Express\_intent\_to\_meet\_or\_negotiate with whom on the 336th day? (GT: John\_Kerry)}\\
    \midrule
    \multirow{7}*{History} & \textbf{Federica\_Mogherini Make\_an\_appeal\_or\_request Iran on the 324th day]}\\
    ~ & \textbf{[Iran Consult Representatives\_(United\_States) on the 323 day]}\\
    ~ & \textbf{[Federica\_Mogherini Discuss\_by\_telephone Mohammad\_Javad\_Zarif on the 289 day]}\\
    ~& \textbf{[Mohammad\_Javad\_Zarif Consult John\_Kerry on the 288 day]}\\
 ~ &\textbf{[Federica\_Mogherini Express\_intent\_to\_engage\_in\_diplomatic\_cooperation\_(such\_as\_policy\_support) Fumio\_Kishida on the 265 day]}\\
    ~ & \textbf{[Federica\_Mogherini Consult Fumio\_Kishida on the 265 day]} \\
    ~ & \textbf{[Fumio\_Kishida Express\_intent\_to\_meet\_or\_negotiate John\_Kerry on the 264 day]} \\
    ~& $\boldsymbol\cdots$ $\boldsymbol\cdots$\\
    \midrule
   Answer & 1. Iran $\xspace$ 2. John\_Kerry $\xspace$ 3. Fumio\_Kishida $\xspace$ $\boldsymbol\cdots$\\
   \midrule
    \multirow{11}*{Explanation} & \textbf{1. Iran: }\\
    ~& Federica\_Mogherini has previously expressed intent to meet or negotiate with Iran on the 324th day. \\
    ~ &It is possible that Federica\_Mogherini may do so again on the 336th day.\\
~& \textbf{2. John\_Kerry: } \\
~& Federica\_Mogherini Discuss\_by\_telephone with Mohammad\_Javad\_Zarif on the 289th day.\\
~& While Mohammad\_Javad\_Zarif have consulted with John\_Kerry on the 288th day. \\
~&It is possible that Federica\_Mogherini may express intent to meet or negotiate with John\_Kerry on the 336th day.\\
~& \textbf{3. Fumio\_Kishida: } \\
~&Federica\_Mogherini has expressed intent to engage in diplomatic cooperation with and consulted Fumio\_Kishida on the 265th day.\\
~&Federica\_Mogherini may express intent to meet or negotiate with Fumio\_Kishida on the 336th day.\\
  ~& $\boldsymbol\cdots$ $\boldsymbol\cdots$\\
            \bottomrule
        \end{tabular}	}
	\caption{An example to show the explainability of the reasoning procedure of \themodel. The explanation of each answer is given by the LLM itself, detailing how the model inferred that answer.}
	\label{explainability}
\end{table*}

\subsection{Analysis of Explainability}
In this section, we aim to explore the explainability of the reasoning procedure of \themodel. To achieve this, we instruct the LLM to explain its inferred answers. An example is shown in Table \ref{explainability}, from which we can observe that the LLM possesses the ability to capture relationships between multi-hop histories. For the inferred answer ``\emph{John\_Kerry}'', we can see from the given explanation that the LLM captures the semantic information of the high-order history chain ``\emph{Federica\_Mogherini$\rightarrow$} \emph{Mohammad\_Javad\_Zarif}\emph{$\rightarrow$John\_Kerry}''. That is, leveraging the LLM's robust semantic understanding, it can provide predictions that are reasonably interpretable to a certain extent. 

\begin{table}[t]
	\centering
  \resizebox{\linewidth}{!}
	{\begin{tabular}{c|ccc}
			\toprule
			 \textbf{Datasets} &  \textbf{ICEWS14} &  \textbf{ICEWS05-15} & \textbf{ICEWS18}\\
			\midrule
			 Ratio of known facts &3.89\% &5.67\% &5.87\%\\
			\bottomrule
	\end{tabular}}
\caption{Known facts checking results.}
	\label{tab:data leakage}
\end{table}

\subsection{Analysis of Data Leakage}
\label{Analysis of Data Leakage}

Due to the absence of publicly disclosed temporal horizons for the pre-trained data of Mixtral-8x7B \citep{jiang2024mixtral}, we consider whether this LLM may have covered some data within ICEWS14, ICEWS18, and ICEWS05-15. To verify this problem, we have a conversation with Mixtral-8x7B to check whether it knows about the facts within the three datasets following \citep{shi2023language}. The ratio of known facts in each dataset is shown in Table \ref{tab:data leakage}, which indicates that Mixtral-8x7B only knows a very limited subset of facts. And the conversations are presented in Appendix \ref{appendix of data leakage}. To avoid the problem of data leakage, we exclude this subset of known facts from the testing set for \themodel reasoning.

\begin{table}[t]
	\centering
  \resizebox{\linewidth}{!}
	{\begin{tabular}{c|cccccc}
			\toprule
			\multirow{2.5}[0]*{\textbf{Model}}  &\multicolumn{2}{c}{\textbf{ICEWS14}} & \multicolumn{2}{c}{\textbf{ICEWS18}}  & \multicolumn{2}{c}{\textbf{ICEWS05-15}} \\
    \cmidrule(lr){2-3} \cmidrule(lr){4-5} \cmidrule(lr){6-7}
    ~ &  Hit@1  & Hit@10 &  Hit@1  & Hit@10 &  Hit@1  & Hit@10\\
			\midrule
	Anon-\themodel &17.00	&48.18		&13.81	&38.39		&23.21	&51.62\\
    \themodel &\textbf{24.2}	&\textbf{51.21}		&\textbf{16.81}	&\textbf{42.68}		&\textbf{25.61}	&\textbf{59.58}\\
			\bottomrule
	\end{tabular}}
\caption{Anonymization experimental results.}
	\label{anon}
\end{table}

\subsection{Analysis on the effect of Prior Knowledge within LLMs}
\label{Prior Knowledge}
In this section, we conduct experiments to investigate how the prior knowledge within LLMs affects the performance of \themodel on TKG prediction. Specifically, we anonymize the TKG data by representing each entity and relation with numerical IDs. The anonymized results of \themodel are presented in Table \ref{anon}, which is denoted as Anon-\themodel. From Table \ref{anon} we can see that the anonymization leads to a certain decline in the performance of \themodel, which indicates that the prior knowledge provides a certain degree of assistance for reasoning. Note that since the possible leakage data is filtered (Section \ref{Analysis of Data Leakage}), we analyze that the usefulness of prior knowledge is primarily attributed to certain static semantic knowledge.

\begin{figure}[t]
	\centering
	\subfloat[ICEWS14s]{\includegraphics[scale=0.46]{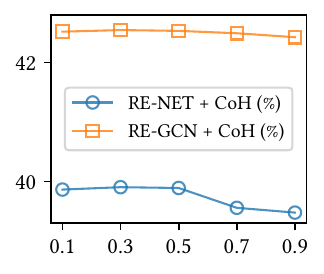}
	}
 \subfloat[ICEWS18]{\includegraphics[scale=0.46]{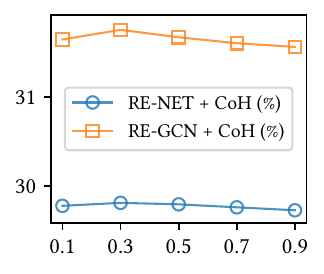}
	}
	\subfloat[ICWS05-15]{\includegraphics[scale=0.46]{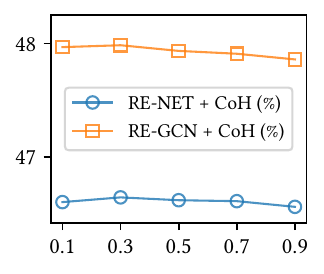}} 
	\caption{Performance of graph-based models plugged with \themodel under different $\alpha$-values in terms of MRR (\%). The x-axis denotes different $\alpha$-values, and the y-axis shows MRR (\%) values.
	}
	\label{hyper-a}
\end{figure}

\begin{figure}[t]
	\centering
	\subfloat[ICEWS14s]{\includegraphics[scale=0.46]{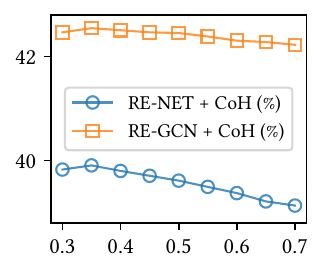}
	}
 \subfloat[ICEWS18]{\includegraphics[scale=0.46]{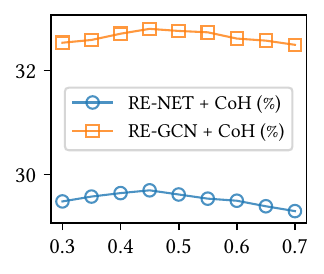}
	}
	\subfloat[ICWS05-15]{\includegraphics[scale=0.46]{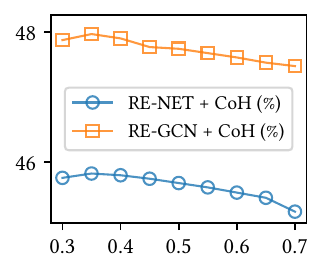}} 
	\caption{Performance of graph-based models plugged with \themodel under different $w$-values in terms of MRR (\%). The x-axis denotes different $w$-values, and the y-axis shows MRR (\%) values.
	}
	\label{hyper-w}
\end{figure}

\subsection{Sensitivity Analysis}
\label{Sensitivity Analysis}
For converting the indexes outputted by LLM into corresponding scores, $\alpha$ determines the score gap for different indexes. we conduct two graph-based models plugged with \themodel when $\alpha$ is in the range of $\{0.1,0.3,0.5,0.7,0.9\}$. The results are shown in Figure \ref{hyper-a}, from which we can see that variations in the value of $\alpha$ within a very narrow range have a minimal impact on the model's performance. 

Moreover, $w$ determines the score weight for fusing the predicted results of graph-based models and \themodel. we conduct two graph-based models plugged with \themodel under various $w$-values. The results are shown in Figure \ref{hyper-w}, from which we can see that in the optimal outcomes, the results of \themodel contribute slightly more to the final score. We analyze the underlying reason leading to the aforementioned observations may be related to the score distribution of the graph-based models.

\section{Related works} 
\paragraph{Temporal Knowledge Graph Forecasting with Supervised Models.} As a carrier of real-world events extracted from news and documents \citep{liu2023enhancing, sun2023noise, liu2024beyond}, TKGs hold significant practical research value. Previous classic methods include GHNN \citep{DBLP:conf/akbc/HanM0GT20} and Know-Evolve \citep{DBLP:conf/icml/TrivediDWS17}, which model the temporal information within TKGs by temporal point process (TTP). And CyGNet \citep{DBLP:conf/aaai/ZhuCFCZ21} proposes a copy-generation mechanism to explore patterns among repetitive histories. As GNNs have shown promise in sequential modeling \citep{liu2017multi}, recently, most supervised models \citep{2020Recurrent, DBLP:conf/sigir/LiJLGGSWC21, DBLP:conf/ijcai/LiS022, zhang2023learning1, liang2023learn, zhang2023learning} for TKG prediction adopt GNNs to capture the structural information within TKGs. In particular, Zhang et al. consider both long-term \citep{wang2024camlo} and short-term information. Based on these, TANGO \citep{han-etal-2021-learning-neural} employs Neural Ordinary Differential Equations to build up continuous temporal information, CENET \citep{xu-etal-2023-cenet} adopts contrastive learning to identify important non-historical entities, MetaTKG \citep{xia2022metatkg} and MetaTKG++ \citep{XIA2024110629} explore the evolution patterns of events with meta-learning, and xERTE \citep{DBLP:conf/iclr/HanCMT21} proposes an explainable model by searching sub-graph in TKGs. Besides, some works \citep{sun-etal-2021-timetraveler, li2021search} search significant paths with reinforcement learning, and Tlogic \citep{liu2022tlogic} extracts paths via temporal logic rules for TKG prediction. 

\paragraph{Temporal Knowledge Graph with Large Language Models (LLMs).} Recently, several works \citep{han2022enhanced, gao2023learning, xu2023pre} have attempted to leverage Pre-trained Language Models (PLMs) on TKG reasoning, which mainly input histories in textual form into PLMs to obtain contextualized knowledge embeddings. Nowadays, with the surge of LLMs, their reasoning capabilities on structural and temporal data are progressively under exploration \citep{jiang2023structgpt, jain2023language, yuan2023back, aghzal2023can, wang2023tram, tan2023towards, xiong2024large}. In the area of TKGs, Ding et al. input relations in textual form into LLMs to generate corresponding descriptions, which are then introduced into embed-based models as a supplement to the semantic information of zero-shot relations \citep{ding2023zero}. And Lee et al. take the first attempt on TKG prediction using LLMs, which is the most closely related work to our paper. They covert TKG prediction into an In-context Learning (ICL) problem, providing LLMs with the first-order histories of the query in textual form to predict the possible answers \citep{lee2023temporal}.

\section{Conclusion}
In this paper, we first analyze the shortcomings and challenges of the existing LLM-based model about how to effectively provide comprehensive high-order historical information for LLM. Then we point out that relying solely on the reasoning capability of LLMs is still limited for TKG prediction. To resolve these issues, we propose \themodel reasoning which achieves effective utilization of high-order histories for LLM. And we design \themodel as plug-and-play, serving to complement and enhance the performance of graph-based models. Extensive experimental results demonstrate the superiority of \themodel, and its effectiveness in enhancing the performance of graph-based models for TKG prediction.

\section{Limitations}
Since \themodel reasoning is conducted in multi-step, LLMs need to be invoked multiple times, resulting in an increased complexity of the inference process. Moreover, we design \themodel as plug-and-play, fusing its answers with predicted results of graph-based models. As this entire process does not involve any training, the fusion weight can only be controlled by the hyper-parameter $w$, making it impossible to achieve adaptive fusion which can automatically learn the weight allocation of the scores obtained from \themodel for different queries. Moving forward, how to design an adaptive fusion strategy that optimally enhances the performance of graph-based models with results of \themodel without compromising efficiency is worth exploring.

\section*{Acknowledgements}
This work is sponsored by the National Key Research and Development Program (2023YFC3305203), and the National Natural Science Foundation of China (NSFC) (Grant 62376265, 62141608, 62206291). 

\bibliography{ref}

\begin{thebibliography}{45}
\expandafter\ifx\csname natexlab\endcsname\relax\def\natexlab#1{#1}\fi

\bibitem[{Aghzal et~al.(2023)Aghzal, Plaku, and Yao}]{aghzal2023can}
Mohamed Aghzal, Erion Plaku, and Ziyu Yao. 2023.
\newblock Can large language models be good path planners? a benchmark and investigation on spatial-temporal reasoning.
\newblock \emph{arXiv preprint arXiv:2310.03249}.

\bibitem[{Boschee et~al.(2015)Boschee, Lautenschlager, O’Brien, Shellman, Starz, and Ward}]{boschee2015icews}
Elizabeth Boschee, Jennifer Lautenschlager, Sean O’Brien, Steve Shellman, James Starz, and Michael Ward. 2015.
\newblock Icews coded event data.
\newblock \emph{Harvard Dataverse}, 12.

\bibitem[{Chen et~al.(2023)Chen, Liao, and Zhao}]{chen2023multi}
Ziyang Chen, Jinzhi Liao, and Xiang Zhao. 2023.
\newblock Multi-granularity temporal question answering over knowledge graphs.
\newblock In \emph{Proceedings of the 61st Annual Meeting of the Association for Computational Linguistics (Volume 1: Long Papers)}, pages 11378--11392.

\bibitem[{Ding et~al.(2023)Ding, Cai, Wu, Ma, Liao, Xiong, and Tresp}]{ding2023zero}
Zifeng Ding, Heling Cai, Jingpei Wu, Yunpu Ma, Ruotong Liao, Bo~Xiong, and Volker Tresp. 2023.
\newblock Zero-shot relational learning on temporal knowledge graphs with large language models.
\newblock \emph{arXiv preprint arXiv:2311.10112}.

\bibitem[{Frantar et~al.(2022)Frantar, Ashkboos, Hoefler, and Alistarh}]{frantar2022gptq}
Elias Frantar, Saleh Ashkboos, Torsten Hoefler, and Dan Alistarh. 2022.
\newblock Gptq: Accurate post-training quantization for generative pre-trained transformers.
\newblock \emph{arXiv preprint arXiv:2210.17323}.

\bibitem[{Gao et~al.(2023)Gao, He, Kan, Han, Qiao, and Li}]{gao2023learning}
Yifu Gao, Yongquan He, Zhigang Kan, Yi~Han, Linbo Qiao, and Dongsheng Li. 2023.
\newblock Learning joint structural and temporal contextualized knowledge embeddings for temporal knowledge graph completion.
\newblock In \emph{Findings of the Association for Computational Linguistics: ACL 2023}, pages 417--430.

\bibitem[{Garc{\'\i}a-Dur{\'a}n et~al.(2018)Garc{\'\i}a-Dur{\'a}n, Duman{\v{c}}i{\'c}, and Niepert}]{garcia2018learning}
Alberto Garc{\'\i}a-Dur{\'a}n, Sebastijan Duman{\v{c}}i{\'c}, and Mathias Niepert. 2018.
\newblock Learning sequence encoders for temporal knowledge graph completion.
\newblock \emph{arXiv preprint arXiv:1809.03202}.

\bibitem[{Han et~al.(2021{\natexlab{a}})Han, Chen, Ma, and Tresp}]{DBLP:conf/iclr/HanCMT21}
Zhen Han, Peng Chen, Yunpu Ma, and Volker Tresp. 2021{\natexlab{a}}.
\newblock Explainable subgraph reasoning for forecasting on temporal knowledge graphs.
\newblock In \emph{ICLR}.

\bibitem[{Han et~al.(2021{\natexlab{b}})Han, Ding, Ma, Gu, and Tresp}]{han-etal-2021-learning-neural}
Zhen Han, Zifeng Ding, Yunpu Ma, Yujia Gu, and Volker Tresp. 2021{\natexlab{b}}.
\newblock Learning neural ordinary equations for forecasting future links on temporal knowledge graphs.
\newblock In \emph{EMNLP}, pages 8352--8364.

\bibitem[{Han et~al.(2022)Han, Liao, Liu, Zhang, Ding, Gu, Koeppl, Schuetze, and Tresp}]{han2022enhanced}
Zhen Han, Ruotong Liao, Beiyan Liu, Yao Zhang, Zifeng Ding, Jindong Gu, Heinz Koeppl, Hinrich Schuetze, and Volker Tresp. 2022.
\newblock Enhanced temporal knowledge embeddings with contextualized language representations.

\bibitem[{Han et~al.(2020)Han, Ma, Wang, G{\"{u}}nnemann, and Tresp}]{DBLP:conf/akbc/HanM0GT20}
Zhen Han, Yunpu Ma, Yuyi Wang, Stephan G{\"{u}}nnemann, and Volker Tresp. 2020.
\newblock Graph hawkes neural network for forecasting on temporal knowledge graphs.
\newblock In \emph{AKBC}.

\bibitem[{Jain et~al.(2023)Jain, Sojitra, Acharya, Saha, Jatowt, and Dandapat}]{jain2023language}
Raghav Jain, Daivik Sojitra, Arkadeep Acharya, Sriparna Saha, Adam Jatowt, and Sandipan Dandapat. 2023.
\newblock Do language models have a common sense regarding time? revisiting temporal commonsense reasoning in the era of large language models.
\newblock In \emph{Proceedings of the 2023 Conference on Empirical Methods in Natural Language Processing}, pages 6750--6774.

\bibitem[{Jiang et~al.(2024)Jiang, Sablayrolles, Roux, Mensch, Savary, Bamford, Chaplot, Casas, Hanna, Bressand et~al.}]{jiang2024mixtral}
Albert~Q Jiang, Alexandre Sablayrolles, Antoine Roux, Arthur Mensch, Blanche Savary, Chris Bamford, Devendra~Singh Chaplot, Diego de~las Casas, Emma~Bou Hanna, Florian Bressand, et~al. 2024.
\newblock Mixtral of experts.
\newblock \emph{arXiv preprint arXiv:2401.04088}.

\bibitem[{Jiang et~al.(2023)Jiang, Zhou, Dong, Ye, Zhao, and Wen}]{jiang2023structgpt}
Jinhao Jiang, Kun Zhou, Zican Dong, Keming Ye, Wayne~Xin Zhao, and Ji-Rong Wen. 2023.
\newblock Structgpt: A general framework for large language model to reason over structured data.
\newblock \emph{arXiv e-prints}, pages arXiv--2305.

\bibitem[{Jin et~al.(2020)Jin, Qu, Jin, and Ren}]{2020Recurrent}
W.~Jin, M.~Qu, X.~Jin, and X.~Ren. 2020.
\newblock Recurrent event network: Autoregressive structure inferenceover temporal knowledge graphs.
\newblock In \emph{EMNLP}, pages 6669--6683.

\bibitem[{Jin et~al.(2019)Jin, Qu, Jin, and Ren}]{jin2019recurrent}
Woojeong Jin, Meng Qu, Xisen Jin, and Xiang Ren. 2019.
\newblock Recurrent event network: Autoregressive structure inference over temporal knowledge graphs.
\newblock \emph{arXiv preprint arXiv:1904.05530}.

\bibitem[{Kwon et~al.(2023)Kwon, Li, Zhuang, Sheng, Zheng, Yu, Gonzalez, Zhang, and Stoica}]{kwon2023efficient}
Woosuk Kwon, Zhuohan Li, Siyuan Zhuang, Ying Sheng, Lianmin Zheng, Cody~Hao Yu, Joseph~E. Gonzalez, Hao Zhang, and Ion Stoica. 2023.
\newblock Efficient memory management for large language model serving with pagedattention.
\newblock In \emph{Proceedings of the ACM SIGOPS 29th Symposium on Operating Systems Principles}.

\bibitem[{Lee et~al.(2023)Lee, Ahrabian, Jin, Morstatter, and Pujara}]{lee2023temporal}
Dong-Ho Lee, Kian Ahrabian, Woojeong Jin, Fred Morstatter, and Jay Pujara. 2023.
\newblock Temporal knowledge graph forecasting without knowledge using in-context learning.
\newblock \emph{arXiv preprint arXiv:2305.10613}.

\bibitem[{Li et~al.(2022)Li, Sun, and Zhao}]{DBLP:conf/ijcai/LiS022}
Yujia Li, Shiliang Sun, and Jing Zhao. 2022.
\newblock Tirgn: Time-guided recurrent graph network with local-global historical patterns for temporal knowledge graph reasoning.
\newblock In \emph{Proceedings of the Thirty-First International Joint Conference on Artificial Intelligence, {IJCAI} 2022, Vienna, Austria, 23-29 July 2022}, pages 2152--2158.

\bibitem[{Li et~al.(2021{\natexlab{a}})Li, Jin, Guan, Li, Guo, Wang, and Cheng}]{li2021search}
Zixuan Li, Xiaolong Jin, Saiping Guan, Wei Li, Jiafeng Guo, Yuanzhuo Wang, and Xueqi Cheng. 2021{\natexlab{a}}.
\newblock Search from history and reason for future: Two-stage reasoning on temporal knowledge graphs.
\newblock In \emph{Proceedings of the 59th Annual Meeting of the Association for Computational Linguistics and the 11th International Joint Conference on Natural Language Processing (Volume 1: Long Papers)}, pages 4732--4743.

\bibitem[{Li et~al.(2021{\natexlab{b}})Li, Jin, Li, Guan, Guo, Shen, Wang, and Cheng}]{DBLP:conf/sigir/LiJLGGSWC21}
Zixuan Li, Xiaolong Jin, Wei Li, Saiping Guan, Jiafeng Guo, Huawei Shen, Yuanzhuo Wang, and Xueqi Cheng. 2021{\natexlab{b}}.
\newblock Temporal knowledge graph reasoning based on evolutional representation learning.
\newblock In \emph{SIGIR}, pages 408--417.

\bibitem[{Liang et~al.(2023)Liang, Meng, Liu, Liu, Tu, Wang, Zhou, and Liu}]{liang2023learn}
Ke~Liang, Lingyuan Meng, Meng Liu, Yue Liu, Wenxuan Tu, Siwei Wang, Sihang Zhou, and Xinwang Liu. 2023.
\newblock Learn from relational correlations and periodic events for temporal knowledge graph reasoning.
\newblock In \emph{Proceedings of the 46th International ACM SIGIR Conference on Research and Development in Information Retrieval}, pages 1559--1568.

\bibitem[{Liu et~al.(2017)Liu, Wu, and Wang}]{liu2017multi}
Qiang Liu, Shu Wu, and Liang Wang. 2017.
\newblock Multi-behavioral sequential prediction with recurrent log-bilinear model.
\newblock \emph{IEEE Transactions on Knowledge and Data Engineering}, 29(6):1254--1267.

\bibitem[{Liu et~al.(2023)Liu, Cheng, Zeng, and Qu}]{liu2023enhancing}
Wanlong Liu, Shaohuan Cheng, Dingyi Zeng, and Hong Qu. 2023.
\newblock Enhancing document-level event argument extraction with contextual clues and role relevance.
\newblock \emph{arXiv preprint arXiv:2310.05991}.

\bibitem[{Liu et~al.(2024)Liu, Zhou, Zeng, Xiao, Cheng, Zhang, Lee, Zhang, and Chen}]{liu2024beyond}
Wanlong Liu, Li~Zhou, Dingyi Zeng, Yichen Xiao, Shaohuan Cheng, Chen Zhang, Grandee Lee, Malu Zhang, and Wenyu Chen. 2024.
\newblock Beyond single-event extraction: Towards efficient document-level multi-event argument extraction.
\newblock \emph{arXiv preprint arXiv:2405.01884}.

\bibitem[{Liu et~al.(2022)Liu, Ma, Hildebrandt, Joblin, and Tresp}]{liu2022tlogic}
Yushan Liu, Yunpu Ma, Marcel Hildebrandt, Mitchell Joblin, and Volker Tresp. 2022.
\newblock Tlogic: Temporal logical rules for explainable link forecasting on temporal knowledge graphs.
\newblock In \emph{Proceedings of the AAAI conference on artificial intelligence}, volume~36, pages 4120--4127.

\bibitem[{Paszke et~al.(2019)Paszke, Gross, Massa, Lerer, Bradbury, Chanan, Killeen, Lin, Gimelshein, Antiga et~al.}]{paszke2019pytorch}
Adam Paszke, Sam Gross, Francisco Massa, Adam Lerer, James Bradbury, Gregory Chanan, Trevor Killeen, Zeming Lin, Natalia Gimelshein, Luca Antiga, et~al. 2019.
\newblock Pytorch: An imperative style, high-performance deep learning library.
\newblock \emph{Advances in neural information processing systems}, 32.

\bibitem[{Shi et~al.(2023)Shi, Xue, Wang, Zhou, Zhang, Zhou, Tan, and Mei}]{shi2023language}
Xiaoming Shi, Siqiao Xue, Kangrui Wang, Fan Zhou, James~Y Zhang, Jun Zhou, Chenhao Tan, and Hongyuan Mei. 2023.
\newblock Language models can improve event prediction by few-shot abductive reasoning.
\newblock \emph{arXiv preprint arXiv:2305.16646}.

\bibitem[{Sun et~al.(2021)Sun, Zhong, Ma, Han, and He}]{sun-etal-2021-timetraveler}
Haohai Sun, Jialun Zhong, Yunpu Ma, Zhen Han, and Kun He. 2021.
\newblock {T}ime{T}raveler: Reinforcement learning for temporal knowledge graph forecasting.
\newblock In \emph{EMNLP}, pages 8306--8319.

\bibitem[{Sun et~al.(2023)Sun, Liu, Wu, Wang, and Wang}]{sun2023noise}
Xin Sun, Qiang Liu, Shu Wu, Zilei Wang, and Liang Wang. 2023.
\newblock Noise-robust semi-supervised learning for distantly supervised relation extraction.
\newblock In \emph{Findings of the Association for Computational Linguistics: EMNLP 2023}, pages 13145--13157.

\bibitem[{Tan et~al.(2023)Tan, Ng, and Bing}]{tan2023towards}
Qingyu Tan, Hwee~Tou Ng, and Lidong Bing. 2023.
\newblock Towards benchmarking and improving the temporal reasoning capability of large language models.
\newblock \emph{arXiv preprint arXiv:2306.08952}.

\bibitem[{Trivedi et~al.(2017)Trivedi, Dai, Wang, and Song}]{DBLP:conf/icml/TrivediDWS17}
Rakshit Trivedi, Hanjun Dai, Yichen Wang, and Le~Song. 2017.
\newblock Know-evolve: Deep temporal reasoning for dynamic knowledge graphs.
\newblock In \emph{ICML}, pages 3462--3471.

\bibitem[{Wang et~al.(2024)Wang, Fu, Wu, Liu, Tan, Huang, Zhang, and Wu}]{wang2024camlo}
Liang Wang, Hao Fu, Shu Wu, Qiang Liu, Xuelei Tan, Fangsheng Huang, Mengdi Zhang, and Wei Wu. 2024.
\newblock Camlo: Cross-attentive multi-view network for long-term origin-destination flow prediction.
\newblock In \emph{Proceedings of the 2024 SIAM International Conference on Data Mining (SDM)}, pages 454--462. SIAM.

\bibitem[{Wang and Zhao(2023)}]{wang2023tram}
Yuqing Wang and Yun Zhao. 2023.
\newblock Tram: Benchmarking temporal reasoning for large language models.
\newblock \emph{arXiv preprint arXiv:2310.00835}.

\bibitem[{Wolf et~al.(2019)Wolf, Debut, Sanh, Chaumond, Delangue, Moi, Cistac, Rault, Louf, Funtowicz et~al.}]{wolf2019huggingface}
Thomas Wolf, Lysandre Debut, Victor Sanh, Julien Chaumond, Clement Delangue, Anthony Moi, Pierric Cistac, Tim Rault, R{\'e}mi Louf, Morgan Funtowicz, et~al. 2019.
\newblock Huggingface's transformers: State-of-the-art natural language processing.
\newblock \emph{arXiv preprint arXiv:1910.03771}.

\bibitem[{Xia et~al.(2024)Xia, Zhang, Liu, Wang, Wu, Zhang, and Wang}]{XIA2024110629}
Yuwei Xia, Mengqi Zhang, Qiang Liu, Liang Wang, Shu Wu, Xiaoyu Zhang, and Liang Wang. 2024.
\newblock Metatkg++: Learning evolving factor enhanced meta-knowledge for temporal knowledge graph reasoning.
\newblock \emph{Pattern Recognition}, page 110629.

\bibitem[{Xia et~al.(2022)Xia, Zhang, Liu, Wu, and Zhang}]{xia2022metatkg}
Yuwei Xia, Mengqi Zhang, Qiang Liu, Shu Wu, and Xiao-Yu Zhang. 2022.
\newblock Metatkg: Learning evolutionary meta-knowledge for temporal knowledge graph reasoning.
\newblock In \emph{EMNLP}, pages 7230--7240.

\bibitem[{Xiang et~al.(2022)Xiang, Cheng, Shang, Zhang, and Liang}]{xiang2022temporal}
Sheng Xiang, Dawei Cheng, Chencheng Shang, Ying Zhang, and Yuqi Liang. 2022.
\newblock Temporal and heterogeneous graph neural network for financial time series prediction.
\newblock In \emph{Proceedings of the 31st ACM International Conference on Information \& Knowledge Management}, pages 3584--3593.

\bibitem[{Xiong et~al.(2024)Xiong, Payani, Kompella, and Fekri}]{xiong2024large}
Siheng Xiong, Ali Payani, Ramana Kompella, and Faramarz Fekri. 2024.
\newblock Large language models can learn temporal reasoning.
\newblock \emph{arXiv preprint arXiv:2401.06853}.

\bibitem[{Xu et~al.(2023{\natexlab{a}})Xu, Liu, Peng, Jia, and Peng}]{xu2023pre}
Wenjie Xu, Ben Liu, Miao Peng, Xu~Jia, and Min Peng. 2023{\natexlab{a}}.
\newblock Pre-trained language model with prompts for temporal knowledge graph completion.
\newblock \emph{arXiv preprint arXiv:2305.07912}.

\bibitem[{Xu et~al.(2023{\natexlab{b}})Xu, Ou, Xu, and Fu}]{xu-etal-2023-cenet}
Yi~Xu, Junjie Ou, Hui Xu, and Luoyi Fu. 2023{\natexlab{b}}.
\newblock Temporal knowledge graph reasoning with historical contrastive learning.
\newblock In \emph{AAAI}.

\bibitem[{Yuan et~al.(2023)Yuan, Xie, Huang, and Ananiadou}]{yuan2023back}
Chenhan Yuan, Qianqian Xie, Jimin Huang, and Sophia Ananiadou. 2023.
\newblock Back to the future: Towards explainable temporal reasoning with large language models.
\newblock \emph{arXiv preprint arXiv:2310.01074}.

\bibitem[{Zhang et~al.(2023{\natexlab{a}})Zhang, Xia, Liu, Wu, and Wang}]{zhang2023learning}
Mengqi Zhang, Yuwei Xia, Qiang Liu, Shu Wu, and Liang Wang. 2023{\natexlab{a}}.
\newblock Learning latent relations for temporal knowledge graph reasoning.
\newblock In \emph{Proceedings of the 61st Annual Meeting of the Association for Computational Linguistics (Volume 1: Long Papers)}, pages 12617--12631.

\bibitem[{Zhang et~al.(2023{\natexlab{b}})Zhang, Xia, Liu, Wu, and Wang}]{zhang2023learning1}
Mengqi Zhang, Yuwei Xia, Qiang Liu, Shu Wu, and Liang Wang. 2023{\natexlab{b}}.
\newblock Learning long-and short-term representations for temporal knowledge graph reasoning.
\newblock In \emph{Proceedings of the ACM Web Conference 2023}, pages 2412--2422.

\bibitem[{Zhu et~al.(2021)Zhu, Chen, Fan, Cheng, and Zhang}]{DBLP:conf/aaai/ZhuCFCZ21}
Cunchao Zhu, Muhao Chen, Changjun Fan, Guangquan Cheng, and Yan Zhang. 2021.
\newblock Learning from history: Modeling temporal knowledge graphs with sequential copy-generation networks.
\newblock In \emph{AAAI}, pages 4732--4740.

\end{thebibliography}

\appendix

\section{Prompt Examples for \themodel Reasoning}
\label{Prompt Example}
Taking the query \emph{(Government\_(Nigeria), Make\_an\_appeal\_or\_request, ?, 340)} as an example, whose ground-truth is ``Member\_of\_the\_Judiciary\_(Nigeria)''. The prompts of \themodel reasoning are shown as follows. 

\clearpage

\begin{tcolorbox}[width=\textwidth, title = {Prompt for Step 1}, center title]
There is a given text consisting of multiple historical events in the form of ``\{id\}:[\{subject\} \{relation\} \{object\} \{time\}];''. And there is a query in the form of: ``\{subject\} \{relation\} \{whom\} time\}?'' If you must infer several \{object\} that you think may be the answer to the given query based on the given historical events, what important historical events do you base your predictions on? Please list the top 30 most important histories and output their \{id\}. \\
\\
Here are the given historical events:\\
0:[Government\_(Nigeria) Engage\_in\_diplomatic\_cooperation with Independent\_Corrupt\_Practices\_Commission on the 339th day]; \\
1:[Government\_(Nigeria) Threaten Education\_(Nigeria) on the 338th day]; \\
2:[Government\_(Nigeria) Criticize\_or\_denounce Boko\_Haram on the 337th day]; \\
3:[Government\_(Nigeria) Threaten Education\_(Nigeria) on the 337th day]; \\
4:[Government\_(Nigeria) Provide\_aid for Citizen\_(Nigeria) on the 337th day];\\
5:[Government\_(Nigeria) Make\_optimistic\_comment on Citizen\_(Nigeria) on the 336th day];\\
6:[Government\_(Nigeria) Use\_conventional\_military\_force to Boko\_Haram on the 335th day];\\
7:[Government\_(Nigeria) Make\_statement to Barack\_Obama on the 333rd day];\\
8:[Government\_(Nigeria) Accuse Boko\_Haram on the 332nd day];\\
9:[Government\_(Nigeria) Express\_intent\_to\_engage\_in\_diplomatic\_cooperation to Citizen\_(Nigeria) on the 332nd day];\\
10:[Government\_(Nigeria) Engage\_in\_diplomatic\_cooperation with Citizen\_(Nigeria) on the 331st day]; \\
$\cdots$ $\cdots$\\
\\
Here is the query:\\
Government\_(Nigeria), Make\_an\_appeal\_or\_request to, whom, on the 340th day?\\
\\
Please only output \{id\} of the historical events that your inferred answers are based on. Note that you must only output no more than 30 \{id\} without any explanation. Please strictly follow the above demands for output.
\tcblower
\textbf{Answer:}\\
\\
0, 1, 2, 5, 7, 8, 9, 10, 11, 14, 17, 21, 24, 25, 26, 27, 28, 30, 31, 32, 33, 34, 36, 37, 38, 39, 45, 49, 50, 56
\end{tcolorbox}

\clearpage

\begin{tcolorbox}[width=\textwidth, title = {Prompt for Step 2}, center title]

You must be able to correctly predict the \{whom\} of the given query from a given text consisting of multiple historical events in the form of ``\{subject\} \{relation\} \{object\} \{time\}'' and the query in the form of ``\{subject\} \{relation\} \{whom\} \{time\}?'' You must output several \{object\} that you think may be the answer to the given query based on the given historical events. Please list all possible \{object\} which may be answers to the query. Please assign each answer a serial number to represent its probability of being the correct answer. Note that answers with a high probability of being correct should be listed first.\\
\\
Here are the given historical events:\\
Government\_(Nigeria), Engage\_in\_diplomatic\_cooperation with, Independent\_Corrupt\_Practices\_Commission, on the 339th day; Independent\_Corrupt\_Practices\_Commission, Arrest\_or\_detain\_or\_charge\_with\_legal\_action to, Citizen\_(Nigeria), on the 308th day;\\
Government\_(Nigeria), Criticize\_or\_denounce, Boko\_Haram, on the 337th day; Boko\_Haram, Use\_conventional\_military\_force to, Citizen\_(Nigeria), on the 336th day;\\
Government\_(Nigeria), Threaten, Education\_(Nigeria), on the 337th day; Education\_(Nigeria), Make\_statement to, Muslim\_(Nigeria), on the 332nd day;\\
Government\_(Nigeria), Make\_optimistic\_comment on, Citizen\_(Nigeria), on the 336th day; Citizen\_(Nigeria), Make\_an\_appeal\_or\_request to, Member\_of\_the\_Judiciary\_(Nigeria), on the 331st day;\\
$\cdots$ $\cdots$\\
\\
Here is the query:\\
Government\_(Nigeria), Make\_an\_appeal\_or\_request to, whom, on the 340th day?\\
\\
lease list all possible \{object\} which may be answers (one per line) without explanations. Note that answers with high probability should be listed first.\\
For example:\\
"""\\
Possible answers:\\
1. XXX\\
2. XXX\\
3. XXX\\
$\cdots$ $\cdots$\\
"""\\
Please strictly follow the above demands for output.\\
\tcblower
\textbf{Answer:}\\
\\
1. Citizen\_(Nigeria)\\ 
2. Education\_(Nigeria)\\
3. Member\_of\_the\_Judiciary\_(Nigeria)
4. Barack\_Obama\\
5. Xi\_Jinping\\
6. Boko\_Haram\\
7. Head\_of\_Government\_(Nigeria)\\
8. Court\_Judge\_(Nigeria)
\end{tcolorbox}

\clearpage

\begin{table}[t]
	\centering
  \resizebox{\linewidth}{!}
	{\begin{tabular}{c|ccc}
			\toprule
			 \textbf{Datasets} &  \textbf{ICEWS14} &  \textbf{ICEWS05-15} & \textbf{ICEWS18}\\
			\midrule
			 $\#$ $\mathcal{E}$ & 6,869     & 10,094 & 23,033 \\
			$\#$ $\mathcal{R}$ & 230    & 251 & 256 \\
			$\#$ Train & 74,845 & 368,868& 373,018  \\
			 $\#$ Valid & 8,514   &46,302  &45,995 \\   
			$\#$ Test  &7,371  &46,159  &49,545 \\
			Time gap  &24 hours  &24 hours &24 hours \\
			\bottomrule
	\end{tabular}}
\caption{The statistics of the datasets.}
	\label{tab:dataset}
\end{table}

\section{Datasets}
\label{Datasets}
In this paper, we utilize three representative TKG datasets for experimental analysis: ICEWS14 \citep{garcia2018learning}, ICEWS18 \citep{jin2019recurrent}, and ICEWS05-15 \citep{garcia2018learning}, all sourced from the Integrated Crisis Early Warning System \citep{boschee2015icews}, documenting events in 2014, 2018, and from 2005 to 2015, respectively. Detailed statistics of the three datasets are shown in Table \ref{tab:dataset}.

\section{Backbones}
\label{backbones}
Since \themodel is plug-and-play, we plug it into several following state-of-the-art TKG reasoning models to evaluate the effectiveness of our proposed model.
\begin{itemize}
    \item \emph{RE-NET} \citep{2020Recurrent} deals with TKGs as KG sequences. RE-NET utilizes the RGCN to capture the structural dependencies of entities and relations within each KG. Then RNN is adopted to associate KGs with different time stamps for capturing the temporal dependencies of entities and relations.
    \item \emph{RE-GCN} \citep{DBLP:conf/sigir/LiJLGGSWC21} proposes a recurrent evolution module based on relational GNNs to obtain embeddings that contain dynamic information for entities and relations. In particular, RE-GCN designs a static module that utilizes the static properties of entities to enrich the embeddings for prediction.
    \item \emph{TiRGN} \citep{DBLP:conf/ijcai/LiS022} utilizes a recurrent graph encoder to capture local temporal dependencies, and designs a history encoder network to capture global temporal dependencies by collecting repeated facts in history. TiRGN fuses the obtained local and global temporal dependencies for final prediction.
\end{itemize}

\section{Implementation Details}
\label{Implementation Details}
In this study, we developed \themodel using the PyTorch framework \citep{paszke2019pytorch}, integrating functionalities from the HuggingFace's Transformers library \citep{wolf2019huggingface} and the efficient LLM inference framework vLLM \citep{kwon2023efficient}. All experiments were executed on an NVIDIA A100 GPU with AMD EPYC 7763 CPU processor. Our experimental setup primarily utilized the \textit{TheBloke/Mixtral-8x7B-Instruct-v0.1-GPTQ} model with a parameter size of 6.07B. This model represents a quantized variant of the \textit{Mixtral-8x7B MoE} model, employing the GPTQ technique \citep{frantar2022gptq} to achieve a balanced compromise between computational speed and model performance. Regarding the generation hyper-parameters, we set the Maximum tokens as 8000, Top-p sampling as 1, and Temperature as 0. 

Moreover, the three graph-based TKG prediction models are all implemented with the most optimal hyper-parameters reported in their corresponding papers. Especially, for evaluating \themodel on reversed testing sets like graph-based models, we manually process relations in datasets into reversed ones. For example, we reverse ``Express intent to cooperate'' into ``Receive intent to cooperate'', and ``threaten'' into ``be threatened''. For the score fusion module, the hyper-parameter $\alpha$ is set to 0.3 for all datasets. And the weights $w$ of ICEWS14s, ICWES18, and ICEWS05-15 are set to 0.35, 0.45, and 0.35, respectively. 

\section{Prompt Examples for Data Leakage Analysis}
\label{appendix of data leakage}
To avoid the possible data leakage issue, we directly ask the LLM Mixtral-8x7B regarding the facts contained in our used datasets, to check whether it knows the knowledge. And we remove all queries whose answer is ``Yes'' from the testing set for all three datasets. A few prompt examples are shown as follows. 

\begin{tcolorbox}[title = {Prompt for Example 1}, center title]
Do you know the fact that United Arab Emirates reduced or broke diplomatic relations with Qatar on 2014-12-04?
\tcblower
\textbf{Answer} No.
\end{tcolorbox}

\begin{tcolorbox}[title = {Prompt for Example 2}, center title]
Do you know the fact that Police (Egypt) used tactics of violent repression against Protesters (Egypt) on 2014-12-02?
\tcblower
\textbf{Answer} Yes.
\end{tcolorbox}

\begin{tcolorbox}[title = {Prompt for Example 3}, center title]
Do you know the fact that Abdullah Abdullah met at a `third' location with Jens Stoltenberg on 2014-12-02?
\tcblower
\textbf{Answer:} No.
\end{tcolorbox}

\end{document}